%% file: 01_neurips_2026.tex
\documentclass{article}

\PassOptionsToPackage{numbers,sort&compress}{natbib}

\usepackage[position, final]{neurips_2026}

\usepackage[utf8]{inputenc} 
\usepackage[T1]{fontenc}    
\usepackage{hyperref}       
\usepackage{url}            
\usepackage{booktabs}       
\usepackage{amsfonts}       
\usepackage{nicefrac}       
\usepackage{microtype}      
\usepackage{xcolor}         
\usepackage[capitalize,noabbrev]{cleveref}
\usepackage{graphicx}
\usepackage{todonotes}
\usepackage{tikz}
\usetikzlibrary{positioning, arrows.meta, shapes.geometric, fit, backgrounds,calc}
\usepackage{amsthm}

\theoremstyle{plain}

\theoremstyle{definition}

\newtheorem{example}{Example}[section]
\theoremstyle{remark}

\title{We Need to Rethink Benchmarking in\\Anomaly Detection}

\author{%
Philipp Röchner$^{*}$\\
Johannes Gutenberg University Mainz\\
University of Southern Denmark\\
\texttt{roechner@uni-mainz.de}\\
\And
Simon Klüttermann$^{*}$\\
TU Dortmund\\
Carnegie Mellon University\\
\texttt{simon.kluettermann@cs.tu-dortmund.de}\\
\And
Kevin Kammler\\
Johannes Gutenberg University Mainz \\
\texttt{k.kammler@uni-mainz.de} \\
\And
Franz Rothlauf \\
Johannes Gutenberg University Mainz \\
\texttt{rothlauf@uni-mainz.de}\\
\And
Emmanuel Müller\\ 
TU Dortmund\\
\texttt{emmanuel.mueller@cs.tu-dortmund.de}\\
\And
Daniel Schlör\thanks{equal contribution}\\ 
University of Würzburg\\
\texttt{schloer@informatik.uni-wuerzburg.de}\\
}

\begin{document}

\maketitle

\begin{abstract}
Despite the continuous proposal of new anomaly detection algorithms and extensive benchmarking efforts, progress seems to stagnate, with only minor performance differences between established baselines and new algorithms. 
We argue in this position paper that limitations in how we evaluate anomaly detection algorithms can hinder progress on specific applications.  
In current benchmarks, a trivial algorithm that only checks for extreme values in individual features performs competitively with state-of-the-art deep learning methods, despite failing on simple cases such as anomalies within an annulus of normal points.
Moreover, existing benchmarks do not adequately reflect the diversity of anomaly detection applications, making it difficult for practitioners to reliably select algorithms for their applications.
Consequently, we need to rethink benchmarking in anomaly detection. 
In our opinion, anomaly detection should be studied using \emph{scenarios} that group applications sharing relevant characteristics, defined through a common taxonomy.
Benchmarking within scenarios enables scenario-specific choices for preprocessing, metrics, and model selection, clarifying which advances transfer across similar applications and providing practitioners with reliable guidance for their specific contexts.
\end{abstract}

\section{Introduction} 

Anomaly detection, also called outlier or novelty detection~\citep{metasurvey}, has many applications, such as detecting attacks on networks~\citep{DBLP:conf/sdm/LazarevicEKOS03}, quality issues in medical data~\citep{rochner2023unsupervised}, or financial fraud~\citep{DBLP:journals/eswa/HilalGY22}.
Although these applications share the property that anomalies are rare and different from the rest of the data~\citep{DBLP:books/sp/Hawkins80}, they assume fundamentally different definitions of anomalies and impose different constraints, such as processing mode, privacy constraints, and label availability. 

Because a single dataset cannot capture all aspects of anomalies, algorithms are evaluated across dataset collections that capture multiple characteristics of anomalies~\citep{DBLP:journals/datamine/CamposZSCMSAH16,han2022adbench}.
This approach separates anomaly detection from other areas of machine learning, such as image classification, where the criteria for a correct prediction are clearer, and fewer datasets are needed for benchmarking~\citep{DBLP:conf/cvpr/DengDSLL009}.

The current evaluation approach for anomaly detection algorithms has limitations: Even though new algorithms are frequently proposed, studies often find that established methods such as $k$-Nearest Neighbors ($k$NN)~\citep{DBLP:conf/sigmod/RamaswamyRS00} and Isolation Forest~\citep{DBLP:conf/iconas/DingF13} perform competitively~\citep{han2022adbench,DBLP:journals/jmlr/BoumanBH24,DBLP:conf/iclr/LivernocheJHR24}.
For example, in the benchmark performed by \citet{han2022adbench} for unsupervised methods, $k$NN ranked best for detecting global anomalies in terms of ROC-AUC; DeepSVDD~\citep{DBLP:conf/icml/RuffGDSVBMK18} and DAGMM~\citep{dagmm}, the only two methods in the benchmark that use neural networks, ranked last and second last, respectively.
Moreover, \citet{DBLP:journals/jmlr/BoumanBH24} identify $k$NN as the best algorithm for datasets with local anomalies and Extended Isolation Forest~\citep{extendedifor} as the best for global anomalies, both outperforming recently published methods.

To illustrate the limitations of current evaluation practices that allow basic methods to produce seemingly convincing results, we discuss the following anomaly detection algorithm \emph{Quantiles}:

\emph{Observations with extreme values in a feature are more likely to be anomalies. 
The anomaly score for each observation is the fraction of features with values less than the $5\%$-quantile or greater than the $95\%$-quantile of the training data.}

\begin{figure}[h!]
\centering
\includegraphics[width=.6\linewidth]{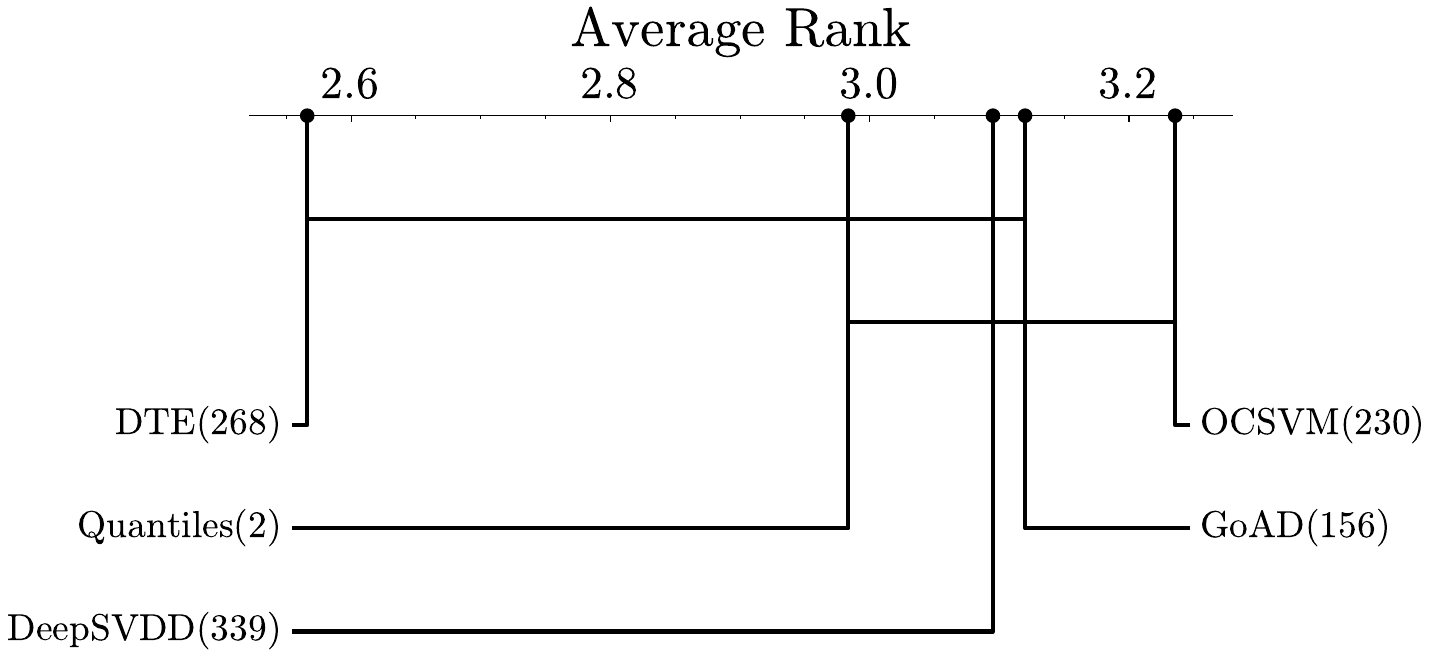}
\caption{Critical difference diagram. 
The numbers in brackets indicate the number of lines of code in each implementation.
They illustrate the brevity of \emph{Quantiles} and do not measure computational or conceptual complexity.
Our simple \emph{Quantiles} algorithm (scoring observations by the fraction of features outside the $5$\%--$95$\% quantile range) 
performs comparably to state-of-the-art methods on ADBench, despite failing on 
simple cases like anomalies within an annulus of normal points. 
This calls into question the practice of demonstrating progress through aggregated performance 
on large dataset collections~\citep{han2022adbench,DBLP:journals/jmlr/BoumanBH24}.}
\label{fig:eye}
\end{figure}

\Cref{fig:eye} shows a critical difference diagram comparing the performance of our \emph{Quantiles} approach with two state-of-the-art approaches (GoAD~\cite{goad} and DTE~\cite{DBLP:conf/iclr/LivernocheJHR24}) and with two established baselines (DeepSVDD~\citep{DBLP:conf/icml/RuffGDSVBMK18} and OCSVM~\cite{moya1993one}) on the ADBench benchmark~\cite{han2022adbench}. 
The training data only contains normal observations, and we use standard hyperparameters.
We then rank the models by ROC-AUC on each benchmark dataset and report their average rank across all datasets.
Algorithms whose performance was not statistically different are connected by horizontal bars.
The number in brackets after each algorithm is the number of lines of code in its implementation, illustrating the brevity of \emph{Quantiles}. 
It does not measure computational or conceptual complexity. 
We follow the PyOD implementations of DeepSVDD and OCSVM, and the authors' implementations of GoAD and DTE.
Our \emph{Quantiles} approach ranks second, slightly ahead of DeepSVDD, GoAD, and OCSVM; only DTE ranks better.
The performance of the \emph{Quantiles} approach was not statistically different from the other methods.
Although our simple competitor seems to perform well, it cannot detect certain types of anomalies. 
Consider, e.g., an anomalous sample in the middle of an annulus of normal points.
These results question the common practice of using a large and diverse collection of datasets for benchmarking to demonstrate improvements in anomaly detection~\citep{han2022adbench,DBLP:journals/jmlr/BoumanBH24}.

Motivated by this observation, our position paper argues that \textbf{we need to rethink benchmarking in anomaly detection.}
We discuss the limitations and potential improvements of current evaluation practices. 
These limitations are a consequence of the inherent challenges of anomaly detection, such as the absence of labels and 
domain-specific definitions of anomalies. 
These challenges, however, do not excuse inadequate benchmarking practices. 
Instead, benchmarks should address these challenges through structured evaluation with clear assumptions and objectives.

In our opinion, there are three areas where benchmarking needs to evolve:
First, anomaly detection should be discussed and evaluated in scenarios that capture the relevant characteristics of multiple applications, such as anomaly type and data modality (see Section~\ref{sc:scenario-specific_benchmarking}).
Second, method development and evaluation should consider all components of anomaly detection pipelines, including preprocessing and hyperparameter selection~(see \cref{sc:components}).
Third, anomaly detection algorithms should be evaluated robustly using scenario-specific objectives, for example, also considering diversity, severity, or fairness (see \cref{sc:meaningful_evaluation}).
Figure~\ref{fig:overview_benchmarking} illustrates how these directions connect.

Although some of these aspects have been explored in supervised machine learning, they are often not directly applicable to anomaly detection. 
Supervised hyperparameter selection approaches, for example, rely on known labels to select \emph{good} hyperparameter configurations. 
They are therefore not directly applicable to unsupervised anomaly detection, where such ground-truth labels are unavailable.

Our goal is to stimulate discussion within the community, leading to more meaningful benchmarks.

\section{Scenario-specific Benchmarking} 
\label{sc:scenario-specific_benchmarking}
\begin{figure*}
    \centering
    \resizebox{\textwidth}{!}{ 
    \input{03_overview_diagram}
    }
    \caption{The proposed scenario-specific benchmarking. 
    Each scenario describes relevant characteristics of multiple anomaly detection applications.
    The chosen scenario determines both the anomaly detection components and the algorithm evaluation.}
    \label{fig:overview_benchmarking}
\end{figure*}

Anomalies are usually defined as patterns in data that deviate from normal behavior.
These patterns originate from the unusual behavior of the underlying generative process~\cite{chandolaAnomalyDetectionSurvey2009,charu-book-outlier-analysis}. 
This perspective requires defining normal behavior and quantifying deviations from it. 
Without explicit and practical definitions of normality, domain experts often determine the semantic boundaries between normal data, noise, and anomalies, tailoring them to their specific application~\cite{charu-book-outlier-analysis}. 
Thus, whether identified through explicit definitions or post hoc analysis, the properties that distinguish anomalies from normal data are application-dependent.

Therefore, new anomaly detection approaches and their evaluations must account for the application context. 
Since anomalies differ widely across applications and data modalities~\cite{foorthuisNatureTypesAnomalies2021}, one cannot expect a one-size-fits-all solution for the \emph{best} anomaly detection algorithm. 
This aligns with the no-free-lunch theorem: different applications have fundamentally different properties, and an algorithm's performance depends on how well its inductive biases match these properties~\cite{calikusNoFreeLunch2020}. 
Therefore, we expect specialized algorithms to perform well in their intended applications.
This finding is also supported by empirical studies~\cite{calikusNoFreeLunch2020,DBLP:journals/jmlr/BoumanBH24}.

Currently, anomaly detection approaches are typically evaluated on benchmark datasets spanning various applications.
Treating results equally across highly dissimilar datasets misses important performance differences: a new algorithm may outperform previous state-of-the-art algorithms on datasets with specific properties. 
This problem is related to Simpson's paradox, in which trends within subgroups can be hidden or even reversed when data are combined~\cite{simpsonInterpretationInteractionContingency1951}. 
Our example from \cref{fig:eye} illustrates that simple baselines perform as well as state-of-the-art algorithms when studying aggregated rankings.

Aggregation across heterogeneous datasets might be common to evaluate anomaly detection algorithms because it provides a unique reference point and guards against cherry-picking when application requirements are unstated. 
We believe, however, that treating all benchmark datasets equally when evaluating algorithms and seeking the best overall performance can obscure or hinder scientific progress in anomaly detection: being \emph{best on the collection} of datasets overlooks the characteristics of specific applications and is not by itself evidence of progress on any particular application. 
\citet{DBLP:journals/tkde/WuK23} observed this ``illusion of progress'' in benchmarks for time series anomaly detection.

Given the number of applications, it is impractical to compare anomaly detection algorithms individually for each application.
Therefore, \textbf{we believe that meaningful comparisons require considering \emph{scenarios}}.
Scenarios describe relevant aspects of different anomaly detection applications, including anomaly type, data modality, and processing mode.
This type of comparison provides a detailed view highlighting improvements in specific scenarios. 

\subsection{Partially Scenario-specific Anomaly Detection Research}

Taxonomies and benchmarks for anomaly detection have been developed that partially adopt a scenario-specific perspective.

\subsubsection{Taxonomies}

The definition of scenarios requires a taxonomy that captures properties, such as anomaly type, latency constraints, or evaluation priorities, that determine algorithm requirements independent of the application.
Most anomaly detection taxonomies~\cite{chandolaAnomalyDetectionSurvey2009,baddarAnomalyDetectionComputer2014,makaniTaxonomyMachineLeaning2018,sebestyenTaxonomyPlatformAnomaly2018,fernandesComprehensiveSurveyNetwork2019,chalapathyDeepLearningAnomaly2019,wittkoppTaxonomyAnomaliesLog2021,pangDeepLearningAnomaly2022,aldayriTaxonomyAnomalyDetection2022,han2022adbench} focus on label availability (supervised, semi-supervised, or unsupervised), algorithm type~\cite{sorboNavigatingMetricMaze2024}, or application aspects~\cite{esteveztapiadorAnomalyDetectionMethods2004,gyanchandaniTaxonomyAnomalyBased2012}. 
For example, the data-centric taxonomy of anomaly characteristics proposed by \citet{foorthuisNatureTypesAnomalies2021} categorizes anomaly detection by dataset, algorithm, and evaluation objective. 

These taxonomies, however, still omit operational constraints of algorithms such as processing mode (real-time versus batch) or the relative importance of false positives versus false negatives. 
Consequently, none yield the scenario definition needed for benchmarking.

\subsubsection{Benchmarks}
\label{sc:benchmarks}

Several studies have highlighted the limitations of aggregating across diverse anomaly detection datasets and have begun shifting toward more nuanced evaluation.

ADBench~\cite{han2022adbench} compares 14 unsupervised detectors across 57 datasets and finds no statistically superior method overall, providing little actionable guidance for practitioners. 
ADBench also performs separate controlled experiments with synthetically generated local, global, dependency, and clustered anomalies, which yields specific recommendations: Local Outlier Factor (LOF)~\citep{lof} is statistically preferable for local anomalies, while $k$NN~\citep{DBLP:conf/sigmod/RamaswamyRS00} performs best for global anomalies. 

Early time-series studies also fix the characteristics of anomalies for evaluation: \citet{DBLP:journals/kais/KeoghLLH07} study the most unusual subsequence of a user-chosen length. 
TSB-AD \cite{liu2024elephant} compares anomaly detection algorithms for time series based on the type of anomaly, such as point-based or sequence-based, and the dimensionality of the data. 

To analyze performance under specific conditions, BMAD \cite{bao2024bmad} organizes six medical imaging datasets around five clinical tasks (brain MRI, liver CT, retinal OCT, chest X-ray, and digital pathology) and two diagnostic goals (image-level and pixel-level anomaly detection).
Their findings differ between the diagnostic goals: memory bank-based algorithms, such as PatchCore~\citep{DBLP:conf/cvpr/RothPZSBG22}, are strong for detecting anomalous samples but produce coarse localization of anomalies in samples. 
Knowledge-distillation methods such as RD4AD~\citep{DBLP:conf/cvpr/DengL22} generally perform better at localization. 
MVTec AD~\citep{DBLP:conf/cvpr/BergmannFSS19} is another image dataset with different texture and object categories and pixel-level ground truth for all anomalies. 
In addition to different objects, semantic anomaly categories, and pixel-level ground truth, \citet{DBLP:conf/eccv/ZouJPZD22} provide images for anomaly detection covering three different domains. 

For anomaly detection in videos, \citet{DBLP:conf/nips/ZhuWRGC24} introduce a multi-environment dataset spanning 14 surveillance domains such as highways, malls, and warehouses, together with distinct evaluation protocols for self-supervised and weakly supervised training. 
The anomalies in the dataset are labeled by semantic anomaly category (e.g., fighting, fire, and object falling) and by domain (e.g., crime, natural hazard, and traffic).
Similar to the semantic anomaly categories, cybersecurity benchmark datasets such as UNSW-NB15, BoT-IoT, ToN-IoT, and CSE-CIC-IDS2018 report performance per attack category (e.g., DoS, reconnaissance, exploits), which enables evaluation within narrower task definitions than aggregate comparisons~\cite{sarhan2021netflow}. 

\citet{macrodata} propose novel datasets for tabular anomaly detection and include limited metadata to distinguish between applications. 
While this is an improvement compared to previous datasets not offering metadata, this metadata is either high-level data characteristics (e.g., "$>20$ features") or limited to LLM-generated source fields ("Origin Science") and thus may not accurately reflect real-world scenarios.

These examples show the feasibility of curating existing datasets and the practical benefit of evaluating methods under well-defined conditions. 
They typically address, however, only a few criteria at a time. 
Sharing a domain label or benchmark collection is not sufficient for compatibility: a dataset labeled ``medical'' or ``cybersecurity'' may differ in supervision regime, required output, or processing mode from other datasets with the same label or in the same collection. 
Moreover, the above studies do not describe how to derive scenarios that cover the relevant requirements of an application. 
A general approach is needed to systematize application requirements, determine which datasets to include in a comparison, identify applicable outputs and metrics, and assess how generalizable the results are.

\subsection{Designing Anomaly Detection Scenarios}
\label{sec:formal_definitions}

We define scenarios and their relationship to anomaly detection applications through \emph{contextual properties}~$\mathcal{P}_{ctx}$ and \emph{structural properties}~$\mathcal{P}_{str}$. 
Structural properties~$\mathcal{P}_{str}$ describe application requirements that substantially affect dataset inclusion, algorithm design, or evaluation protocol. 
These properties include data modality, anomaly type, or processing mode.
Contextual properties~$\mathcal{P}_{ctx}$ describe the deployment setting of an anomaly detection application, such as the application domain (e.g., healthcare, finance) or the application goal being pursued.
They are not directly relevant for algorithm design or evaluation, but they serve as input for domain experts to identify the structural properties~$\mathcal{P}_{str}$ based on real-world requirements. 
For instance, patient monitoring in intensive care is a contextual property: domain experts translate this property into structural properties, such as real-time processing and high recall to avoid missing critical events. 
These properties directly constrain algorithm and evaluation choices.

We define a scenario as a group of applications that share the same value for every structural property included in the scenario declaration, regardless of differences in their contextual properties~$\mathcal{P}_{ctx}$. 
The scenario declaration must include all application requirements that substantially affect dataset inclusion, algorithm design, or evaluation protocol.
Applications that differ in any such requirement cannot belong to the same scenario.
This decouples benchmarking from specific applications: results from a comparison within a scenario provide relevant evidence for every application that satisfies the scenario declaration.

Identifying the relevant structural properties requires following established methodologies for taxonomy development~\cite{DBLP:journals/ejis/NickersonVM13} in collaboration with domain experts, ensuring that scenarios reflect real-world requirements. 
Properties can be split or merged iteratively, adding granularity when heterogeneous applications share a scenario, and consolidating when scenarios become too narrow to cover multiple similar real-world applications.

\begin{table}[h]
\centering
\caption{Exemplary structural properties~$\mathcal{P}_{str}$ for anomaly detection scenarios. The structural properties and their values need to be iteratively extended or refined.}
\label{tab:struc_prop}
\begin{tabular}{ll}
\toprule
\textbf{Structural property~$\mathcal{P}_{str}$} & \textbf{Possible values} \\
\midrule
Anomaly type       & global, local, clustered, contextual/sequential \\
Anomaly ratio & very low ($<$1\%), low (1--5\%), moderate ($>$5\%) \\
Data modality      & tabular, time series, image/spatial, text, graph \\
Data dimensionality & low ($<$10), medium (10--100), high ($>$100) \\
Supervision        & unsupervised, semi-supervised, supervised \\
Processing mode & real-time/online, batch/offline \\
Evaluation priority  & precision, recall, specificity \\
Operational constraints & robustness, fairness, diversity
\\ 
\bottomrule
\end{tabular}
\end{table}

Table~\ref{tab:struc_prop} provides an extensible set of candidate structural properties.
Each scenario declaration includes all relevant entries and refines their values where needed. 
It also adds any missing relevant properties.
The selection of structural properties~$\mathcal{P}_{str}$ and their values determines the number of scenarios and the number of applications per scenario.

We emphasize that we are not proposing to evaluate all combinations of structural property values.
Instead, we rely on two guardrails: First, a scenario is formalized only when it covers at least a minimum number of applications.
Second, scenarios in which algorithmic performance is indistinguishable should be investigated more closely and, if appropriate, merged.

The structural properties that define a scenario also determine its appropriate evaluation objectives, which we explore in \cref{sc:objective}.

\subsection{Examples of Anomaly Detection Scenarios}
\label{sec:scenario_examples}

The following examples illustrate how applications with different contextual properties~$\mathcal{P}_{ctx}$ can share a scenario when they satisfy the same structural properties~$\mathcal{P}_{str}$. 

\begin{example}[One-class Detection and Localization in Images]
\label{examples:scenario1}
    This scenario captures applications requiring detection and pixel-level localization of anomalies in single 2D images.
    Contexts include visual inspection of manufactured components and textured materials~\cite{DBLP:conf/cvpr/BergmannFSS19,DBLP:conf/eccv/ZouJPZD22}.
    Structurally, anomalies are spatially localized, training uses only normal target-task images, and processing is batch/offline.
    The required outputs are an image-level anomaly score and a pixel-level anomaly map.
    Evaluation prioritizes precision and recall for anomalous pixels under strong class imbalance, emphasizing pixel-level PR-AUC for localization and reporting image-level ROC-AUC separately for detection.
\end{example}

\begin{example}[Batch Detection of Diverse Novelties]
\label{examples:scenario2}
    This scenario groups applications where batch processing is sufficient.
    Contexts may include quality control for medical databases~\cite{rochner2023unsupervised} and scientific discovery in astronomy~\cite{activepersonalisation}, provided their data modality and remaining declared requirements match.
    The goal is to identify unknown behaviors from unlabeled data using unsupervised methods.
    The evaluation objective shifts from pure detection accuracy to diversity, aiming to retrieve a wide range of root causes rather than repeated instances of a single anomaly type.
\end{example}

\begin{example}[Detection of Adversarial Anomalies]
\label{examples:scenario3}
    The final scenario involves adversaries.
    Contexts may include financial fraud detection~\cite{fraudapl} and network intrusion detection~\cite{auto_appl_httptraffic}, provided their data modality, supervision regime, processing mode, and remaining declared requirements match.
    Structurally, we consider cases where anomalies are intentionally hidden to mask their anomalous nature, appearing normal in isolation but deviating from their local neighborhoods.
    Due to the high volume of data, the difficulty of detecting hidden anomalies, and the cost of investigating flagged cases, algorithms must prioritize high-precision results.
\end{example}

\section{Scenario-specific Anomaly Detection Components} 
\label{sc:components}

Researchers usually propose end-to-end anomaly detection pipelines consisting of distinct \emph{components}, such as data preprocessing, model 
selection, and ensemble construction~\cite{usualAD1,usualAD2,usualAD3,DBLP:conf/icml/RuffGDSVBMK18,goad}. 
Because components interact, it is often unclear whether observed improvements arise from a new algorithm or from other pipeline choices, and whether a component that helps in one pipeline generalizes to others. 
This ambiguity is compounded by the absence of a shared application context: without fixing the structural properties~$\mathcal{P}_{str}$ of the application, it is considerably more difficult to systematically select relevant components, assess which interactions matter, or determine whether a finding transfers beyond the original pipeline.

Therefore, \textbf{we advocate for research on scenario-specific anomaly detection components.}
Instead of evaluating only the overall performance of end-to-end systems, researchers should analyze components individually, making contributions explicit and reusable across 
pipelines. 
\citet{jiangADGymDesignChoices2023} support this, showing that a modular perspective improves the performance of deep one-class anomaly detection over an end-to-end approach.
Scenarios simplify this modular analysis in practice by informing decisions about each component: they constrain which preprocessing choices are viable (\cref{sc:preprocessing}), help inform model selection (\cref{sc:model_selection}), and narrow the set of admissible algorithms for ensemble creation  (\cref{sc:ensembles}). 

\subsection{Data Preprocessing}
\label{sc:preprocessing}

The choice of data preprocessing can substantially affect anomaly detection performance~\citep{DBLP:journals/datamine/Kandanaarachchi20}. 
Most preprocessing steps, however, are designed without considering anomaly-relevant structure, and common choices can hide anomalies. 
Dimensionality reduction, such as PCA, captures dominant variance but discards low-variance dimensions that may distinguish anomalies from normal observations.
Data transformations such as min-max scaling and z-score 
standardization suppress the extreme-value structure that anomaly detectors often
rely on. 
Robust alternatives such as median-and-MAD standardization can preserve this structure. 

We therefore argue that preprocessing steps should be designed specifically for anomaly detection, preserving the structure that distinguishes anomalies from normal data.
Without an explicit application context, however, this is difficult because the same normalization that preserves anomaly-relevant signal in one setting can hide it in another.

Scenarios resolve this ambiguity by formalizing structural properties as explicit preprocessing constraints, limiting the preprocessing choices. 
For example, a scenario with a declared latency budget rules out any feature-extraction step that exceeds that budget, even if the step would improve detection quality.
Similarly, a scenario in which anomalies manifest as extreme-tail events rules out any normalization that compresses those tails.

\subsection{Model Selection}
\label{sc:model_selection}

Researchers constantly propose new anomaly detection algorithms~\cite{lof,ecod,kde,aean,copod,nf,gmm,dagmm,gan,DBLP:conf/icml/RuffGDSVBMK18,DBLP:conf/iclr/LivernocheJHR24}. 
Choosing a \emph{good} one with appropriate hyperparameters is often crucial for 
detection performance~\citep{ding2022hyperparameter}, yet difficult without labeled data: unlike supervised learning, there is typically no ground truth to evaluate candidate models.
Some model selection approaches exist~\cite{zhaoModelselection,zhaoMetaOD,
hypexTimeseriesHP,flaml,marques2020internal} but provide limited 
benefit~\cite{zhaoNeedFor}. 

We believe that a key challenge for model selection is the broad and diverse nature of anomaly detection.
To guide algorithm and hyperparameter selection, algorithmic characteristics should align with scenario requirements, including anomaly type, data characteristics, and processing constraints. 
First, scenarios constrain the class of admissible algorithms: for example, scenarios specifying local anomalies admit LOF and rule out methods designed for global density estimation. 
Second, scenarios convert model selection into a transfer problem with a well-founded inductive basis: 
performance on one dataset within a scenario provides evidence for performance on another, because the shared structural properties~$\mathcal{P}_{str}$, such as anomaly type, data modality, and processing constraints, determine the algorithmic requirements for all datasets in that scenario. 
While similar performance cannot be guaranteed, this is a substantially stronger basis for model selection than the unsupervised alternative, which provides no foundation for generalization across datasets.

\subsection{Ensembles}
\label{sc:ensembles}

Ensembles that combine different anomaly detection methods often outperform individual models~\cite{plaincombine,ijcnnensemble}. 
Best practices for designing ensembles, however, remain unclear because the large space of possible combinations makes systematic study impractical. 
This problem is particularly challenging for unsupervised anomaly detection because labeled data to evaluate model combinations is missing.
Consequently, ensemble improvements currently rely on heuristics and personal experience~\cite{aggarwal_outlier_ensembles_book}. 

A modular approach combined with scenario-specific guidance transforms ensemble design from an exponential search guided by personal experience into a smaller, constrained selection problem. 
Scenarios narrow the design space for ensembles: researchers select components that satisfy the scenario's requirements rather than searching through all possible combinations.
For example, the scenario's computational budget determines which ensembles are viable: real-time scenarios may exclude ensembles if the total component costs exceed the available latency budget, whereas batch scenarios might not have any declared runtime budget. 
When combining multiple models is feasible, the scenario specifies which additional requirements the models must meet: a scenario containing both local and global anomalies requires models for each anomaly type, for example, LOF for local and Isolation Forest for global anomalies~\cite{han2022adbench}. 

\section{Scenario-specific Evaluation} 
\label{sc:meaningful_evaluation}

Evaluating anomaly detection algorithms raises the following questions: 
What are the application's objectives? 
Which metrics best reflect them? 
How can we reliably estimate them given the small number of anomalies in the dataset? 
And which structural assumptions of a scenario enable theoretical analysis to complement empirical evaluation?

The answers to these questions depend on the scenario and its structural properties~$\mathcal{P}_{str}$. 
Therefore, \textbf{anomaly detection algorithms should be evaluated on a scenario-by-scenario basis}, with metrics that reflect scenario objectives and protocols that account for scenario constraints.

\subsection{Evaluation Objectives}
\label{sc:objective}

ROC-AUC and PR-AUC are commonly used metrics for anomaly detection.
These metrics, however, assume error costs that may not match the scenario's priorities~\citep{charu-book-outlier-analysis,mcdermott2025closerlookaurocauprc}. 
PR-AUC implicitly penalizes false positives in high-confidence regions, making it suited for scenarios where false alarms are costly (e.g., intrusion detection). 
ROC-AUC evaluates global class separation, making it preferable when missing anomalies is the primary risk (e.g., medical diagnosis)~\citep{mcdermott2025closerlookaurocauprc}. 

Even with the appropriate choice among these metrics, however, both remain rank-based and may overlook other properties, such as diversity, severity, or fairness.
For instance, in identifying quality issues in a database, prioritizing fewer but more diverse and severe anomalies is often more valuable than detecting many similar, insignificant ones (see~\cref{sec:scenario_examples}).
Diversity metrics, well-studied in recommender systems~\citep{DBLP:journals/kbs/KunaverP17}, could be adapted for this purpose.
Other factors relevant to specific domains include runtime (see Section~\ref{sec:scenario_examples}), robustness to noise in healthcare~\cite{adRobustApplication}, fairness in business screening~\cite{adFairApplication}, and interpretability in engineering~\cite{adInterpretableApplication}.

\textbf{We argue that the anomaly detection community should study and adopt evaluation metrics that reflect scenario-specific objectives.}  
Researchers should therefore examine whether established metrics truly capture the objectives of the scenario, select the most appropriate ones, and, where needed, complement them with additional measures for previously overlooked or inadequately captured properties.
Moreover, new relevant metrics can stimulate the development of algorithms specifically designed to optimize for these properties.

\subsection{Evaluation Protocols}
\label{sc:protocols}

In addition to selecting evaluation metrics, the structural properties~$\mathcal{P}_{str}$ of a scenario help determine which evaluation protocol can produce reliable estimates of those metrics.
Many algorithms, such as Isolation Forests~\citep{DBLP:journals/tkdd/LiuTZ12}, incorporate random components, and datasets often consist of single samples from complex generative processes for both anomalies and normal observations.

Evaluations based on random subsets, such as train-test or $k$-fold cross-validation, are unreliable in scenarios with few observations and a low anomaly ratio as a structural property (see \cref{sec:formal_definitions}): subsets may miss entire anomaly types, as observed in some ADBench datasets~\cite{han2022adbench}.
Unlike supervised learning on datasets with a balanced distribution of labels, in which both classes have sufficient samples for reliable estimation, anomaly detection scenarios inherently involve rare events, which limit statistical power and complicate evaluation protocols.

Such scenarios require protocols such as bootstrapping~\cite{bootstrappingOld,bootstrappingNew} and statistical evaluation of performance variability across multiple initializations, for example, using confidence intervals~\citep{boyd2013area}.
Small sample sizes, however, can still complicate statistical interpretations~\citep{boyd2013area,efron1979bootstrap}, and developing robust evaluation protocols for low-anomaly-count settings remains an open challenge.

\subsection{Theoretical Analysis and Controlled Experiments}
\label{sc:theoretical_analysis}

Anomaly detection algorithms are rarely analyzed theoretically and using controlled experiments, despite such analysis could provide insights beyond those offered by limited benchmark datasets.
This likely originates from the broad nature of the field, as theoretical analysis and controlled experiments require well-defined assumptions.
For example, a density-based algorithm might detect anomalies in one-class settings because low-density samples are more likely to be anomalies. 
However, in an unsupervised setting where clustered anomalies form high-density regions~\cite{bachelorsthesisCriticKnn}, it can fail. 
Such implicit assumptions make theoretical analysis and controlled experiments challenging.

Scenarios address this by incorporating explicit structural properties (see \cref{sc:scenario-specific_benchmarking}), enabling findings from theoretical analysis and controlled experiments to apply across algorithms designed for the same scenario constraints.
For instance, \citet{ttt} highlight an algorithm's failure to detect rare anomalies, using controlled experiments to identify the minimum frequency required for successful detection.

\section{Call to Action}

Current benchmarking overlooks fundamental differences of anomaly detection applications. 
Although specialized benchmarks for anomaly detection exist (see \Cref{sc:benchmarks}), they remain limited to a few criteria, and a systematic way to obtain these criteria is missing.
We therefore argue for systematically defining anomaly detection scenarios and using them for benchmarking, where \emph{scenarios} capture the relevant characteristics of multiple applications. 

Community collaboration can support wider adoption of scenario-specific benchmarking through the following steps:

\begin{enumerate}
    \item \textbf{Taxonomy and scenario definition:} Researchers organize workshops to develop an initial anomaly detection taxonomy and derive scenarios and evaluation objectives from it. 
    \item \textbf{Dataset curation:} Dataset creators collaborate with domain experts to curate scenario-specific datasets with suitable evaluation objectives and standardized metadata, including contextual and structural properties.
    These datasets are published in dedicated dataset journals and on conference tracks. 
    \item \textbf{Scenario-specific research:} Researchers develop scenario-specific algorithms for data preprocessing, anomaly detection, and model selection. 
    Conference organizers and journal editors encourage the publication of results by scenario in special tracks rather than endorsing unnecessary and uninformative aggregation across all datasets. 
    \item \textbf{Scenario application:} Practitioners use the scenario taxonomy to select algorithms and share feedback through competitions and repositories.\footnote{At \url{https://go.c1a.eu/benchad}, researchers and practitioners can share and discuss scenarios, applications, and datasets.}
\end{enumerate}
The taxonomy defined in step one will be refined iteratively based on the insights from steps two through four.

We believe that scenario-specific benchmarking will enable researchers to make measurable progress on well-defined problems and provide practitioners with reliable guidance for selecting algorithms for their applications.

\section{Alternative Views}

\subsection{Number versus Relevance of Benchmarking Datasets}

\paragraph{Alternative view:} 
Scenario-specific evaluation risks fragmenting the field. 
Isn't it better to evaluate on a broader range of diverse datasets, as in current benchmark practice, rather than focusing on a few carefully curated ones?
Large benchmark collections provide a common reference point for comparing anomaly detection methods, enabling researchers to understand how their methods generalize to other contexts. 
This approach focuses on methodological advancements and enables faster development within established frameworks. 
In contrast, an overemphasis on scenario-specific evaluation can lead to a fragmented collection of niche benchmarks that hinder cross-domain comparisons.

\paragraph{Our view:} While collections of benchmark datasets are useful, aggregation across heterogeneous datasets with incompatible requirements fails to capture the diversity of anomaly detection: 
anomaly definitions and their contexts vary across applications and are not solely data-driven. 
Rather, the distinction between anomalous and normal observations is typically made by domain experts. 
Therefore, meaningful evaluation requires scenarios that make these definitions explicit. 
While we acknowledge concerns about fragmentation, the current aggregation of heterogeneous datasets produces comparability that is broad but shallow in depth. 
Scenario-specific evaluation trades cross-scenario breadth for within-scenario depth, where comparisons are more meaningful. 
This depth enables hypothesis-driven research: rather than pursuing an unrealistic overall \emph{best} solution on heterogeneous benchmarks, researchers can define specific claims and design algorithms to address them within well-defined scenarios. 

\subsection{Anomaly Detection Pipelines versus Components}

\paragraph{Alternative view:} Individual anomaly detection components cannot detect anomalies; only complete pipelines of components can do so.
Consequently, researchers typically design anomaly detection pipelines to maximize end-to-end performance. They rarely expect components to perform well in other anomaly detection pipelines.
Benchmarking studies evaluating component combinations would exponentially increase effort. 

\paragraph{Our view:} While anomaly detection components are often optimized for specific pipelines, it is reasonable to expect that some will perform well in novel combinations.
By exploring these combinations, we can identify components that enhance performance, generalizing beyond the original setup. 
Proposing individual anomaly detection components rather than whole pipelines allows researchers to clarify their contributions and makes advances more reusable since well-studied modular components are more likely to be adopted in future research.
Additionally, designing algorithms based on a single pipeline using ground-truth labels can lead to overfitting, which may hinder generalization to similar scenarios.

\subsection{Few Standard versus Many Scenario-specific Evaluation Objectives}

\paragraph{Alternative view:} Although scenario-specific objectives can be useful, having more objectives makes it easier to identify those in which a proposed algorithm will perform well. 
When every paper reports superiority on different niche objectives, it becomes difficult for practitioners to meaningfully compare methods and make selections. 
Furthermore, achieving consensus on evaluation objectives may be unrealistic.
Sticking to a few standard evaluation objectives provides reliable evaluations, even if they are suboptimal in certain scenarios.
That being said, a large set of evaluation objectives already exists, allowing researchers to select appropriate metrics based on their research questions rather than relying exclusively on standard measures.

\paragraph{Our view:} We agree that defining objectives solely to demonstrate an algorithm's superiority over baselines is undesirable. 
Instead, objectives should be determined by the scenario rather than the algorithm itself. 
The key question is which relevant properties should be measured in a given scenario?
Scenario-specific benchmarking facilitates this: rather than seeking consensus on general objectives for anomaly detection, the community only needs to reach an agreement within each scenario, where the objectives and constraints are clearer. 
Importantly, this means leveraging established objectives and metrics where applicable, and fixing agreed-upon evaluation protocols for each scenario. 
Alongside these comparisons, common objectives from established benchmarks can still be reported to maintain comparability with prior work.

\section*{Acknowledgments and Disclosures}

This work was partly supported by the Lamarr-Institute for ML and AI and the Bavarian Ministry of Economic Affairs, Regional Development and Energy through the BRACE-LLM project (grant no. DIK-2407-0004 // DIK0726/03).
The Linux HPC cluster at TU Dortmund University, a project of the German Research Foundation, provided the computing power. 
The authors declare no competing interests. 

\bibliographystyle{unsrtnat} 
\bibliography{02_ref_simon,02_references}

\end{document}

%% file: 03_overview_diagram.tex
\def\sdis{0.25cm}
\def\mdis{0.5cm}
\def\ldis{1.4cm}

{
\footnotesize
\begin{tikzpicture}[
    arrow/.style={->, >=latex, thick},
    components/.style={draw, fill=blue!10, rectangle, thick},
    scenarios/.style={draw, fill=green!10, rectangle, thick},
    evaluation/.style={draw, fill=orange!10, rectangle, thick},
    app/.style={draw, fill=yellow!10, rectangle, thick},
    container/.style={draw, rounded corners, thick, fill=gray!5},
    node distance=0.5cm
]

\tikzset{
    every node/.style={
        minimum width=2cm,
        text width=2.4cm,
        minimum height=0.8cm,
        align=center
    }
}

\node (preprocessing) [components] {Preprocessing (\cref{sc:preprocessing})};
\node (model) [components, right= of preprocessing] {Model selection (\cref{sc:model_selection})};
\node (ensemble) [components, right= of model] {Ensemble (\cref{sc:ensembles})};

\node (components) [draw, rounded corners, fit=(preprocessing) (model) (ensemble), inner sep=0.3cm, label=below:{Components (\cref{sc:components})}] {};

\node (etc) [scenarios, above=\ldis of ensemble] {...};
\node (anomaly_type) [scenarios, left=of etc] {Anomaly type};
\node (data_type) [scenarios, above=of anomaly_type] {Data modality};
\node (processing) [scenarios, right=of data_type] {Processing mode};

\node (scenarios) [draw, rounded corners, fit=(anomaly_type) (data_type) (processing) (etc), inner sep=0.3cm, label=above:{Scenarios (\cref{sc:scenario-specific_benchmarking})}] {};

\draw[arrow] (preprocessing.east) -- (model.west);
\draw[arrow] (model.east) -- (ensemble.west);

\draw[arrow] (scenarios.south) -- (scenarios.south |- components.north);

\node (objective) [evaluation, right=\ldis of processing] {Evaluation\\objectives\\(\cref{sc:objective})};
\node (protocols) [evaluation, below= of objective] {Evaluation\\protocols\\(\cref{sc:protocols})};
\node (theoretical) [evaluation, below= of protocols] {Theoretical analysis\\controlled exps.\\(\cref{sc:theoretical_analysis})};

\node (evaluation) [draw, rounded corners, fit=(objective) (theoretical), inner sep=0.3cm, label=above:{Evaluation (\cref{sc:meaningful_evaluation})}] {};

\draw[arrow] (scenarios.east) -- (evaluation.west |- scenarios.east);
\draw[arrow] (components.east) -- (evaluation.west |- components.east);

\node (app1) [app, left=\mdis of scenarios] {Application ...};
\node (app2) [app, above= of app1] {Application 1};
\node (app3) [app, below= of app1] {Application $n$};

\draw[arrow] (app1.east) -- (scenarios.west |- app1.east);
\draw[arrow] (app2.east) -- (scenarios.west |- app2.east);
\draw[arrow] (app3.east) -- (scenarios.west |- app3.east);

\end{tikzpicture}
}